\pdfoutput=1

\documentclass[11pt]{article}

\usepackage[]{acl}

\usepackage{times}
\usepackage{latexsym}

\usepackage[T1]{fontenc}

\usepackage[utf8]{inputenc}

\usepackage{inconsolata}
\usepackage{microtype}

\usepackage{inconsolata}
\usepackage{hyperref}
\usepackage{url}
\usepackage{hyperref}       
\usepackage{url}            
\usepackage{booktabs}       
\usepackage{amsfonts}       
\usepackage{nicefrac}       
\usepackage{microtype}      
\usepackage{xcolor}         
\usepackage{url}            
\usepackage{booktabs}       
\usepackage{amsfonts}       
\usepackage{nicefrac}       
\usepackage{microtype}      
\usepackage{subcaption}
\usepackage{amsmath,amsfonts,amssymb}
\usepackage{breqn}
\usepackage{tabularx}
\usepackage{multirow}
\usepackage{graphicx}
\usepackage{color}
\usepackage{tcolorbox}
\usepackage{CJK}
\usepackage{adjustbox}
\usepackage{xcolor}
\usepackage{colortbl}
\usepackage{multicol}
\usepackage{vwcol} 
\newsavebox{\algleft}
\newsavebox{\algright}
\usepackage{bm}
\usepackage{booktabs}
\usepackage{amsmath,stackengine}
\usepackage[linesnumbered,ruled,vlined]{algorithm2e}
\usepackage{times}
\usepackage{latexsym}
\usepackage{bbm}
\usepackage[utf8]{inputenc} 
\usepackage[T1]{fontenc}    
\usepackage{hyperref}       
\usepackage{url}            
\usepackage{booktabs}       
\usepackage{amsfonts}       
\usepackage{nicefrac}       
\usepackage{microtype}      
\usepackage{xcolor}         
\usepackage{amssymb}
\usepackage{pifont}
\usepackage{bbm}
\usepackage{breqn}
\usepackage{longtable}
\usepackage{algpseudocode}

\title{Bridging the Preference Gap between Retrievers and LLMs}

\author{Zixuan Ke$^{2}$\thanks{~~The work was done during internship at Google Research.}, 
~Weize Kong$^{1}$,
~Cheng Li$^{1}$, 
~Mingyang Zhang$^{1}$,
~Qiaozhu Mei$^{3}$\thanks{~~The work was done as a visiting researcher at Google Research.} 
~and Michael Bendersky$^{1}$ 
\\ 
$^1$Google Research\\
$^2$University of Illinois at Chicago\\
$^3$University of Michigan\\
$^1$\texttt{\{weize,chgli,mingyang,bemike\}@google.com} \\
$^2$\texttt{zke4@uic.edu}\\
$^3$\texttt{qmei@umich.edu} \\
}

\begin{document}
\maketitle
\begin{abstract}
Large Language Models (LLMs) have demonstrated superior results across a wide range of tasks, and Retrieval-augmented Generation (RAG) is an effective way to enhance the performance by locating relevant information and placing it into the context window of the LLM. However, the relationship between retrievers and LLMs in a RAG is still under-investigated. Most existing work treats the retriever and the LLM as independent components and leaves a gap between retrieving human-``friendly'' information and assembling a LLM-``friendly'' context. In this work, we examine a novel bridge mechanism. We validate the ranking and selection assumptions of retrievers in the context of RAG and propose a framework that chains together supervised and reinforcement learning to train a bridge model that optimizes the connection between the retriever and the LLM. Empirical results demonstrate the effectiveness of our method in both question-answering and personalized generation tasks.
\end{abstract}




\section{Introduction}
\label{sec.intro}



Large language models (LLMs) such as GPT-4 \cite{openai2023gpt4} and PaLM 2 \cite{DBLP:journals/corr/abs-2305-10403}, have demonstrated impressive performance on a wide variety of tasks. 
Retrieval-augmented generation (RAG), which retrieves knowledge items from an external data source and puts it into the context window of LLMs, has produced significantly enhanced results in many NLP tasks \cite{Khandelwal2020Generalization,borgeaud2022improving,izacard2022few,yasunaga2023retrieval}.  

\begin{figure}[t!]
\centering
\includegraphics[width=\columnwidth]{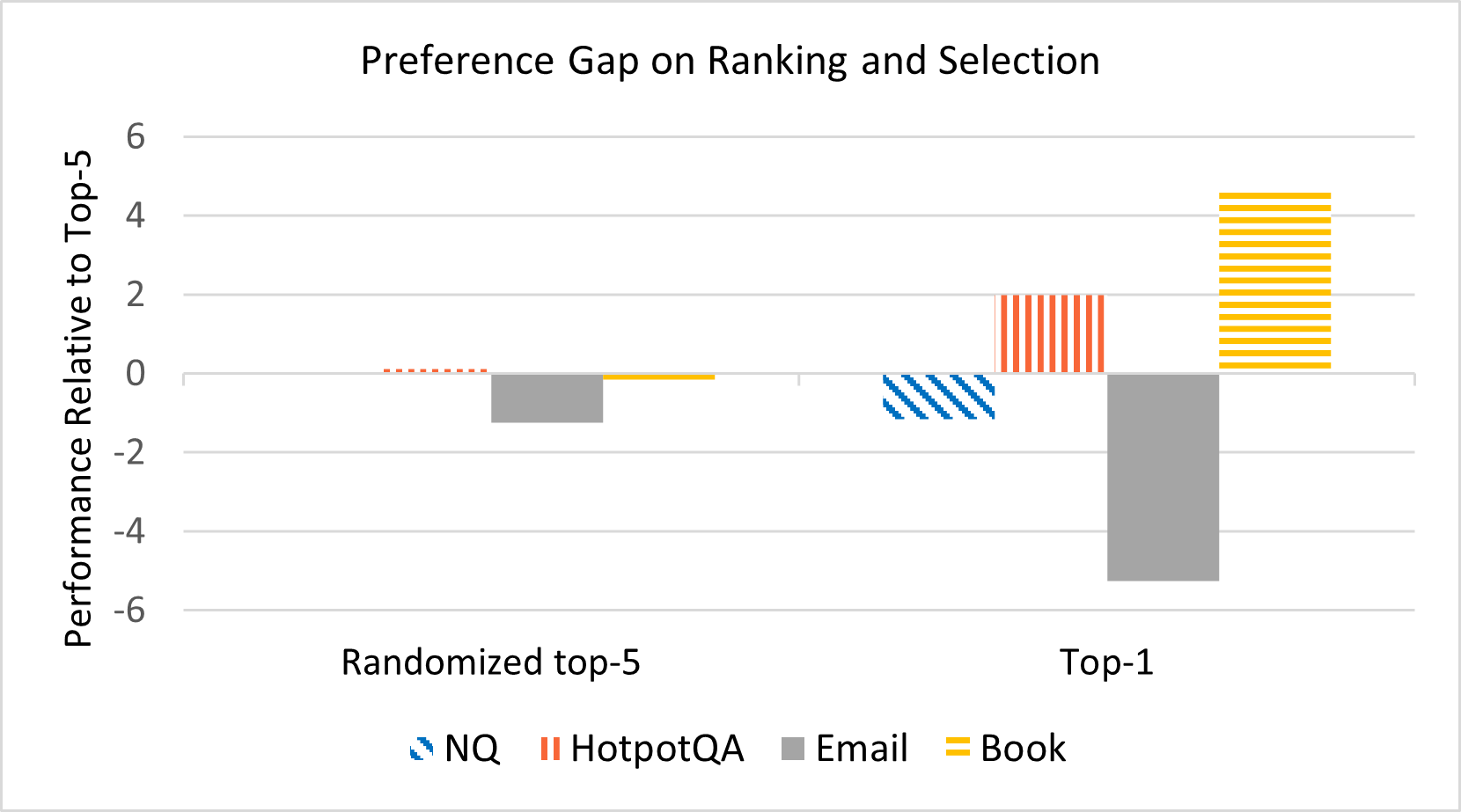}
\caption{We observe a preference gap when alternating the ranking and selection of information in RAG. Experiments are conducted with retrieving passages using GTR~\cite{ni2021large} and using top K of them as additional context for a frozen Palm2-S LLM. 
Different colors indicate different datasets (detailed in Sec.~\ref{sec:datasets}) and the Y-axis shows the relative percentage. 
Alternating the selection (Top-1) of information significantly affects (either positively or negatively) the LLM's performance, while randomizing the ranking of multiple selected items (Top-5) does not have a comparable impact (the metrics are detailed in Sec.~\ref{sec.parameter}). Note the impact on NQ is even too small to be visible.
}
\label{fig.preference_gap}
\end{figure}

However, most works on RAG study retrievers and LLMs \textbf{separately}.
On one hand, most retrievers are designed to be \textbf{human-friendly}, usually based on the \textbf{general belief} in classic information retrieval literature that ranking is paramount, as humans typically read from top to bottom \cite{robertson1977probability}. On the other hand, LLMs exhibit preferences different from humans and yield accurate results only when the information in the prompt aligns with these preferences. This discrepancy leads to sub-optimal design in current RAG systems, a phenomenon we term \textbf{preference gap}. This gap manifests in various aspects. For example, the general belief in \textbf{ranking} may not align with LLM's preferences due to the self-attention mechanism of Transformers, which can focus on any token regardless of its position.
Another aspect is \textbf{selection}; while humans can easily disregard irrelevant information in a context, it has been shown that LLMs are highly sensitive to irrelevant content \cite{shi2023large}. There likely exist more aspects that further diverge the LLM's preference from that of humans, e.g., \textbf{repetition}. Repeated information is generally considered detrimental for retrieval systems \cite{xia2017adapting}, but it may be useful to LLMs for weighting the relevance of context items.

We empirically investigate this preference gap, focusing specifically on \textbf{ranking} and \textbf{selection}. As shown in Fig.~\ref{fig.preference_gap}, when we randomize the ordering of top-5 retrieved items (in our case, passages), the performance of RAG only varies by around 1\% \footnote{We note that this finding is different from \cite{liu2023lost}, where a ``loss in the middle'' phenomenon is observed, meaning that the LLM is better at using the relevant information at the beginning or end of its input context. This is probably due to different numbers of retrieved items. \citet{liu2023lost} used 20 documents, whereas we use 5 passages. Regardless, both findings indicate that there exists a preference gap between the retrievers and LLMs, which we aim to bridge.}. However, the variation exceeds 5\% when the LLM is only presented with the top-1 passage under each order (therefore it encounters different \textit{selections} of information). This indicates the general belief in ranking does not always apply to LLMs, and that the selection of information could be more crucial. This finding confirms the existence of the preference gap between retrievers and LLMs, and it is critical to \textbf{bridge this preference gap} to enhance the performance of RAG. To the best of our knowledge, this is a novel insight that may guide future designs of RAG systems.



Existing work has tried to finetune the LLMs to align with the retriever or adjust the retriever to align with the LLM. However, finetuning LLMs, especially at the scale of GPT-4 or Palm 2, is often expensive. Similarly, it is difficult to adjust production-level retrievers such as Google or Bing. Even when the retriever is adjustable, existing efforts often focus on re-ranking the retrieved results and fail to address other aspects of preference such as selection or repetition. 
Instead, we propose a novel and practical framework called \textbf{BGM} (\textbf{B}ridging the \textbf{G}ap between retrievers and LL\textbf{M}s), which keeps the retriever and LLM fixed and trains a bridge model in between. The bridge model aims to transform the retrieved information into a format that LLMs prefers and can effectively work with. 

Without loss of generality, we structure the bridge model as a \textbf{sequence-to-sequence (seq2seq)} model, which allows dynamically selecting items, re-ranking them, and potentially broader operations like repeating some of them in the retrieval-augmented prompt. Training such a bridge model is challenging as there are usually no ground truth labels on ideal \textbf{item sequences} for retrieval augmentation. Existing work has tried to derive supervisory signals for ranking from RAG's downstream task performance, such as perplexity distillation ~\cite{izacard2022few}. Nevertheless, these methods only provide pointwise supervisory signals for each item independently. Directly applying the same idea to obtain sequential supervision is infeasible, since it would require feeding all possible item sequences into the LLM to obtain perplexity. We developed a greedy search approach to solve this problem (Sec. \ref{sec.sl}). Moreover, we find sequential supervision can be too sparse to effectively train such a seq2seq model (Table~\ref{tab.abalation_SL_only}). To address this issue, we employed reinforcement learning (RL) on the SL trained bridge model, regarding the downstream task performance as the reward and the bridge model as a policy model. Chaining SL and RL provides increased supervision from the downstream task. It also offers the flexibility to explore more advanced strategies, such as repetition, in forming the optimal passage sequence.



Our experiments reveal that BGM can enhance the performance of various downstream tasks, such as Question Answering (QA) and personalized generation, across a spectrum of datasets, from public QA and amazon reviews to private email conversations. Notably, the modified passages retrieved by BGM surpass the performance of strong retrievers and baseline reranking models. This underscores the significance and promise of the ``bridge'' approach in the realm of RAG. In summary, our contributions can be summarized as follows:

\begin{itemize}
    \vspace{-2mm}
    \item We empirically establish the existence of the preference gap between retrievers and LLMs, and introduce BGM to address this gap.
    \vspace{-2mm}
    \item  We propose a seq2seq bridge model to jointly accomplish reranking and selection, adapting the retrieved information to be LLM-friendly. We employ a SL and RL training scheme to optimize this adaptation process.
    \vspace{-2mm}
    \item We evaluate BGM with diverse tasks, including QA and text generation, with publicly available and personalized datasets. The evaluation underscores the effectiveness of BGM in bridging the preference gap and improving RAG performance in downstream tasks.
\end{itemize}

\begin{figure}[t!]
\centering
\includegraphics[width=\columnwidth]{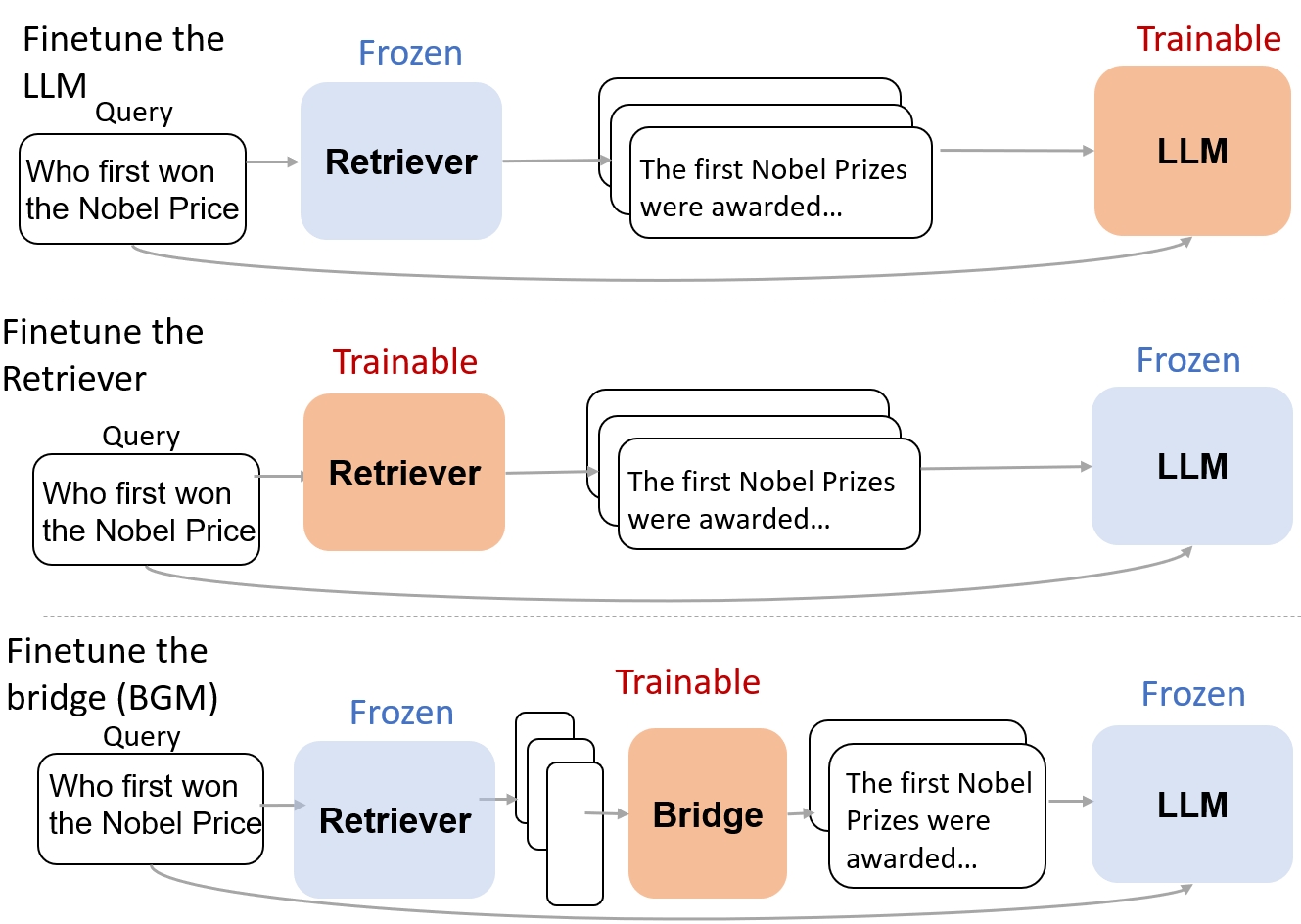}
 \vspace{-5mm}
\caption{Differing from previous RAGs that update LLMs, retrievers, or both, BGM connects a frozen LLM and a frozen retriever through a lightweight bridge model which adapts the retrieved information to the LLM's preference. This makes BGM applicable to ``large'' LMs and off-the-shelf retrievers.
}
\label{fig.comparison}
\vspace{-6mm}
\end{figure}

\section{Related Work}
\label{sec.relate}


\textbf{Retrieval-augmented Generation (RAG).}
Augmenting LLMs with relevant information retrieved from various knowledge sources is proven effective across numerous NLP tasks, including language modeling \cite{borgeaud2022improving,Khandelwal2020Generalization,shi2023replug}, question answering \cite{lewis2020retrieval,izacard2022few,de2023glimmer,de2023pre,shi2023replug,guu2020retrieval,izacard2020leveraging,xu2023recomp}, fact versification \cite{lewis2020retrieval} and text generation \cite{lewis2020retrieval}. Specifically, RAG utilizes input as a query and comprises two main components: (1) a \textbf{retriever} retrieves a set of items from a side corpus. Particular items may vary across different tasks, including documents, passages, or even tokens. In this study, we focus on retrieving passages; and (2) a \textbf{LLM} incorporates the retrieved items, as additional information in the input context, and makes final predictions.

A fundamental question in this process arises regarding the disparate preferences between LLMs and retrievers, as LLMs performing optimally only when their preferences are satisfied. Bridging the preference gap is crucial. Depending on which components are subject to updates, this challenge can be categorized into three families.

\textbf{Finetuning retrievers and LLMs jointly.} This is the most widely used setting of RAG \cite{izacard2022few,Khandelwal2020Generalization,wu2022memorizing,guu2020retrieval,lewis2020retrieval}. However, most prior work is based on relative small LMs ($<$ 1B). For example, Altas~\cite{izacard2022few} finetunes LLM (T5~\cite{raffel2020exploring}) and retriever (Contriever~\cite{izacard2021unsupervised}) jointly by leveraging the LLM to provide supervisory signal to train the retriever. RAG~\cite{lewis2020retrieval} uses a tunnable query encoder and DPR~\cite{karpukhin2020dense} as retriever, BART as LLM, and design an end-to-end schema to train the query encoder and the LLM.

\textbf{Finetuning LLMs only.} Updating retrievers is not always desirable as it is costly and requires the document index to be periodically updated. To bridge the preference gap, it is also possible to only update the LLMs. FiD \cite{izacard2020leveraging} takes the retrieved documents and query as input, finetunes the LLM to adapt to the external information. Similarly,  Lummen~\cite{de2023pre} and Glimmer~\cite{de2023glimmer} improve FiD via adding reranker and pre-encoding memeory.

\textbf{Finetuning retrievers only.} Although above systems have shown improves results, they are not always applicable in practice. Many LLMs (especially larger ones) are only available as black-box APIs. 
Given this constraint, a natural approach is to only update the retrievers to ensure they can retrieve passages that are more compatible with LLMs. REPLUG~\cite{shi2023replug} adapts a similar idea as Atlas but fix the LM. RECOMP~\cite{xu2023recomp} trains a compressors to summarize the retrieved document from retriever. However, this family of models is incapable of performing any sample-level selection and can only choose top passages by setting a fixed threshold.

Unlike the existing approaches, BGM does not fine-tune the LLM or the retriever and instead employs a bridge model in between (Fig.~\ref{fig.comparison}). \newline



\noindent\textbf{RL for information retrieval.} Before the LLM era, RL has been used in information retrieval (IR) \cite{xia2017adapting,wei2017reinforcement,zeng2018multi,xu2020reinforcement}. The core approach was to frame the IR problem as a Markov Decision Process and apply an RL algorithm to solve it. Typically, an IR task would be structured to determine which document to select for a ranking position, using ranking metrics such as DCG as the reward. 
None of these existing studies explore the application of RL in the context of RAG.

In the LLM era, RL has been used in query rewriting for retrieval \cite{wu2021conqrr,nogueira2017task,adolphs2021boosting}, where a black-box retriever is assumed. This is a different problem from RAG. \citet{bacciu2023rraml} suggest using RL to fine-tune retriever in the RAG context  in their opinion paper, not supported by experiments. Their work does not recognize the importance of bridging the gap between retrievers and LLMs, nor does it specify what the bridge model should be.
 



\section{Problem Formulation}
\label{sec.problem_formulate}

\textbf{Retriever. }Given an input $x$, the retriever aims to retrieve a ranked list of passages from a corpus $D = \{d_i\}_{i=1}^m$ that are relevant to $x$. In this work, we assume a typical scenario of employing a frozen dense retriever. Typically, a dual encoder architecture is applied, where an encoder is used to encode both the input context $x$ and the passage $d$. Specifically, the encoder maps each passage to an embedding $\bm{E}(d)$. The similarity between input and passage embedding is computed by their cosine similarity,
\begin{equation}
    s(d,x) = \text{cos}(\bm{E}(d), \bm{E}(x)).
\end{equation}

The top-k passages that have the highest similarity scores when compared with the input $x$ are retrieved in this step, 
\begin{equation}
\label{eq.topk_retrieve}
    (d^{\text{retr.}}_{j})_{j=1}^{k} = \text{Top-K}(\{s(d,x)\}_{i=1}^m).
\end{equation}

\textbf{Bridge Model for RAG. } 
The retrieved top-K passages provide richer information about the original input/query $x$ to help the LLM to make a better prediction on downstream tasks. A bridge model $\bm{B}$ adapts the retrieved passages to a sequence of passages that is LLM-friendly. As mentioned in the Sec. \ref{sec.intro}, the bridge model is a seq2seq model. It takes all the retrieved passages $(d^{\text{retr.}}_{j})_{j=1}^{k}$ as well as the query $x$ as input, and outputs the adapted passages $(d^{\text{bdr.}}_{j})_{j=1}^{n}$,
\begin{equation}
    (d^{\text{bdr.}}_{j})_{j=1}^{n} = \bm{B}\left(x,(d^{\text{retr.}}_{j})_{j=1}^{k}\right).
    \label{eq.bridge model}
\end{equation}

This formulation is general enough as the seq2seq model automatically considers \textit{ranking} by generating the next token based on the preceding one, \textit{selection} by placing the end-of-sentence token in the appropriate position, and \textit{repetion} by generating the same passage ID (as explained in the following paragraph). Note that $n$ may be smaller or larger than $k$ due to selection and repetition. 


Before concatenating the query and passages as bridge model's input, we prepend each passage with a unique sentinel token as its passage ID, e.g.,
$[\text{query}][\text{id}_1]d^{\text{retr.}}_{1}[\text{id}_2]d^{\text{retr.}}_{2}$. In this way, the model only needs to generate the passage IDs instead of the actual passage content, which is much more efficient and avoids making unfaithful changes to the retrieved passages. We then convert the obtained passage IDs to the corresponding passages for downstream processing.


\textbf{Retrieval-augmented generation with bridge.} We concatenate adapted passages from the bridge model, $(d^{\text{bdr.}}_{j})_{j=1}^{n}$, with the input $x$, and fed the resulting long sequence into the LLM as context to obtain the output for downstream tasks. 




\begin{figure*}[t!]
\centering
\includegraphics[width=\textwidth]{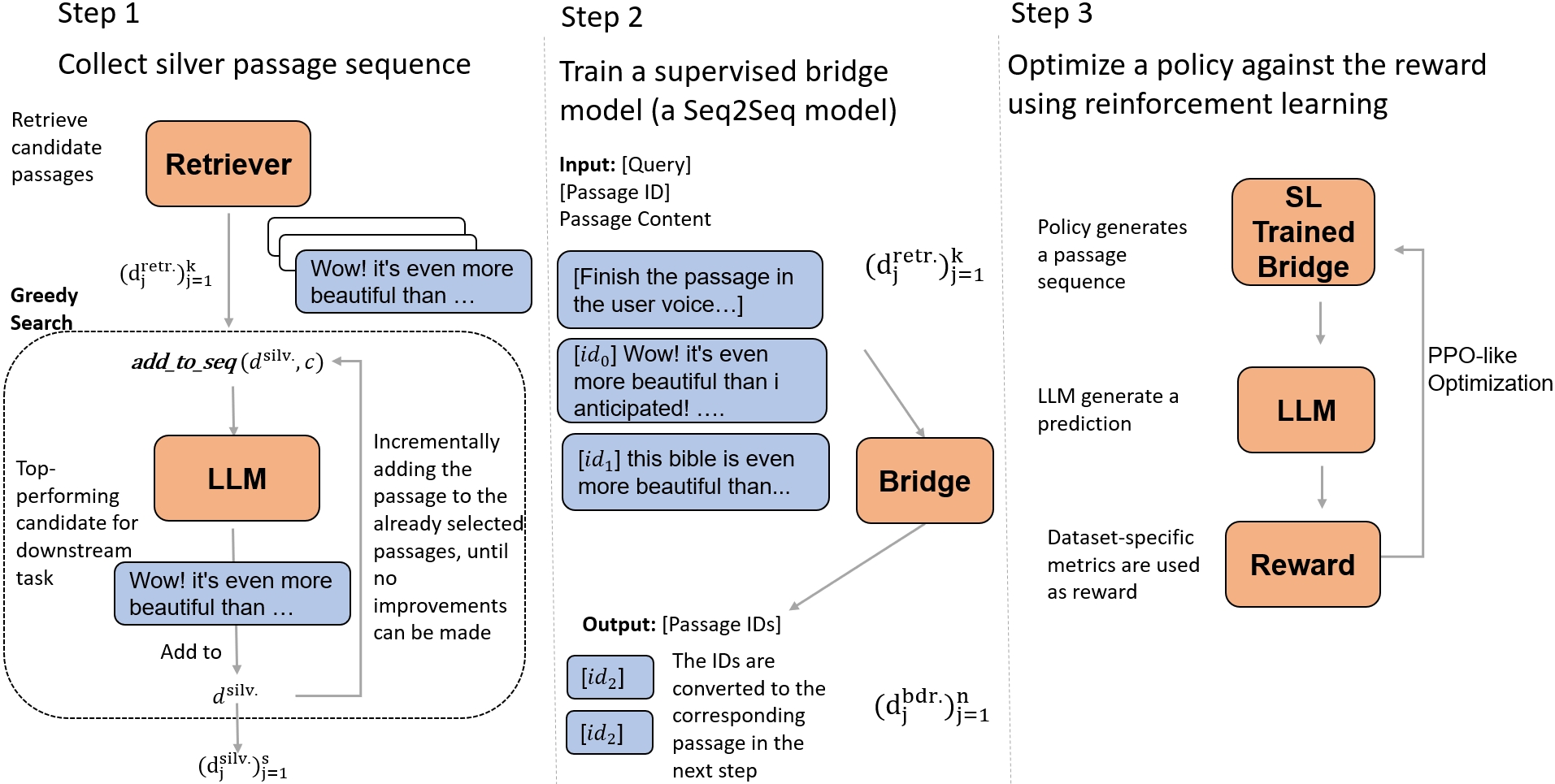}
\caption{Illustrating the training of BGM: \textbf{Step 1:} First, we prepare the silver passage sequence for supervised learning (SL) through a greedy search on the retrieved passages. \textbf{Step 2:} These silver passage sequences are then used for the supervised training of the bridge model. The bridge model is a seq2seq model that takes the query and passages with prepended passage IDs as input and outputs the passage IDs. \textbf{Step 3:} Finally, the SL-trained bridge model is treated as a policy model and is further trained using reinforcement learning.
}
\label{fig.overview}
\end{figure*}

\section{Training the Bridge Model}

In Eq.~\ref{eq.bridge model}, we format the bridge model as seq2seq model. As mentioned in Sec.~\ref{sec.intro}, its training is challenging due to the lack of supervision for \textbf{passage sequences}, the infeasibility of directly applying existing methods that provide only point-wise supervision, and the sparsity of sequential supervision. For effective training, we propose to chain supervised learning (SL) and reinforcement learning (RL), where SL aims to reduce the search space of RL and provides a reasonably good initial model that does ranking and selection. RL aims to optimize the policy model, i.e., bridge model, for the downstream task. Fig.~\ref{fig.overview} shows an overview.

\begin{algorithm}[!]
\caption{Synthesis SPS}\label{alog.silver}
\KwIn{$(d^{\text{retr.}}_{j})_{j=1}^{n}$, $R(\cdot)$} 
\KwOut{$(d_j^{\text{silv.}})_{j=1}^s$ } 
$d^{\text{silv.}} \gets ()$\;
$R^{\text{silv.}} \gets R(d^{\text{silv.}})$\;
$R^{\text{best}} \gets -\infty$\;
\While{true}{
    \For{$c \leftarrow d^{\text{retr.}}_1$ \KwTo $d^{\text{retr.}}_n$ where $c \not\in d^{\text{silv.}}$}
    {
    $d' \gets \mathtt{add\_to\_seq}(d^{\text{silv.}}, c)$\;
    $R^{\text{cur}} \gets R(d')$\;
    \If{$R^{\text{cur}} > R^{\text{best}}$}
    {
        $c^{\text{best}} \gets c$\;
        $R^{\text{best}} \gets R^{\text{cur}}$\;
    }
    }
    \eIf{$R^{\text{best}}> R^{\text{silv.}}$}{
       $d^{\text{silv.}} \gets \mathtt{add\_to\_seq}(d^{\text{silv.}}, c^{\text{best}})$\;
       $R^{\text{silv}} \gets R^{\text{best}}$\;
       $R^{\text{best}} \gets -\infty$\;
       }{
       \textbf{break}\;
      }  
    
}
\end{algorithm}
\vspace{-6mm}

\subsection{Supervised Learning}
\label{sec.sl}

To conduct SL, the ground-truth passage sequence is required for each query. Existing approaches, which focus on obtaining the relevance score of each passage (see Section \ref{sec.intro}), are not applicable as we need to determine which \textit{combination of passages} is most effective for the downstream tasks. To address this, we propose to synthesis silver passage sequence (SPS) by selecting only the useful passages. This is done by \textbf{greedy search} that incrementally selects the next passage that can maximize the downstream task performance.


\textbf{Synthesising SPS using greedy search.}
We denote the downstream task performance when using a given passage sequence for RAG as $R(\cdot)$. For the edge case where the passage sequence is empty, $R(\varnothing)$ simply denotes the task performance without retrieval augmentation (i.e., no passage used). We start from an empty SPS $d^{\text{silv.}}=\varnothing$, and iteratively add the next best candidate passage to the sequence, measured based on the resulted task performance $R(d^{\text{silv.}})$. We stop until no improvement can be made to $R(d^{\text{silv.}})$. Algorithm \ref{alog.silver} shows the pseudo-code for synthesising SPS. The training with the SPS is achieved by applying cross-entropy loss.

\subsection{Reinforcement Learning}
\label{sec.rl}

Although SL can already help training the bridge model, it is still ineffective -- we observe using SL alone results in mixed performance (see Table \ref{tab.abalation_SL_only}). This is attributed to sparse supervision (see Sec. \ref{sec.intro}) and the lack of end-to-end training on downstream results.

To address these issues, we apply RL to continue the training of the bridge model. 
In SL, we only consider permutations or deletions in the SPS, while RL can accommodate more complex manipulations that an optimal passage sequence might require, such as repetition. Additionally, RL provides enhanced supervision beyond the silver sequences through the reward from sampled passage sequences. Using the performance of downstream task as the reward, the bridge model is trained in an end-to-end manner.
Specifically, our task can be formulated as an RL problem --
\textbf{Reward} is the performance of the downstream task, usually measured based on certain ground-truth labels, e.g., Extract Match or BLEU. 
\textbf{Policy model} is the bridge model that need to be trained. 
\textbf{Action space} is restricted to passage IDs (Sec. \ref{sec.problem_formulate}) as we are interested in organizing rather than revising the retrieved passages. In training, the reward objective can be optimized by any off-the-shelf RL algorithm, e.g., proximal policy optimization (PPO).




\section{Experiments}

\subsection{Datasets and Baselines}
\label{sec:datasets}

\textbf{Datasets}. We consider four datasets, ranging from popular QA datasets to personalized generation datasets. We also include one dataset that contains private email conversations (Avocado Email), which is unlikely to be included in the LLM's pre-training datasets. This will further help us investigate the effectiveness of our proposed BGM model, as the LLM will have to rely on the retrieved passages. The summary of statistics is given in Table~\ref{tab.dataset}. 

\textbf{Open-domain QA}. We conduct evaluations on two open-domain QA datasets: \textbf{Natural Questions (NQ)} \cite{kwiatkowski-etal-2019-natural} and \textbf{HotpotQA} \cite{DBLP:conf/emnlp/Yang0ZBCSM18}. They both consist of questions, answers collected from Wikipedia and the Web. HotpotQA is a multi-hop QA dataset which requires findings and reasoning over multiple passages to answer the question. The candidate passages are retrieved from Wikipedia pages.

\begin{figure}[t!]
\centering
\includegraphics[width=0.6\columnwidth]{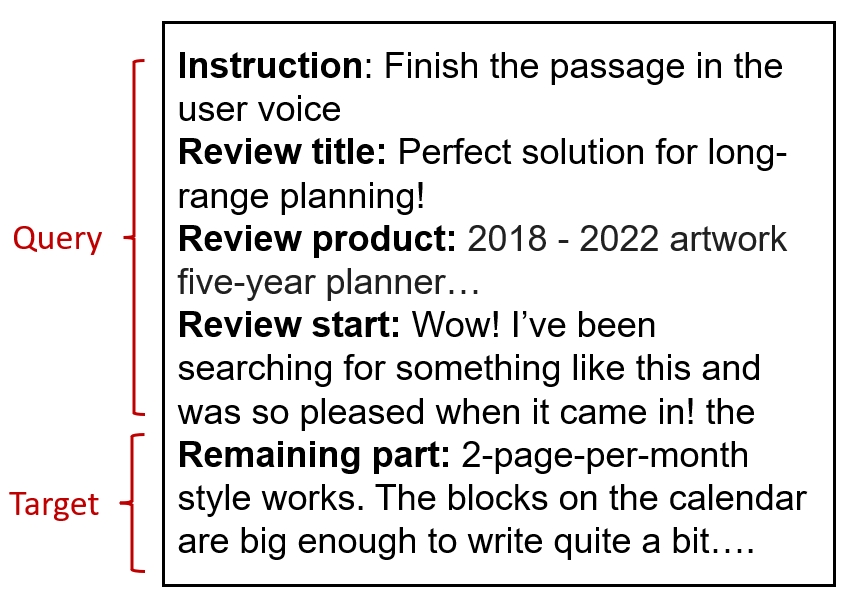}
 \vspace{-3mm}
\caption{An example from the Book dataset. The query consists of the instruction, title, product (there will be no product information for Email) and document start. The goal of the task is to generate the remaining part of the document. 
}
\label{fig.example}
\vspace{-3mm}
\end{figure}

\textbf{Personalized Generation. }We follow \cite{li2023teach} to construct the personalized generation datasets. In this task, the query includes the start and necessary properties (e.g., title) of a document authored by a user. The objective is to complete the document as if it were finished by the same user. The candidate passages for this task are retrieved from documents previously authored by this user.
This includes datasets from two domains: \textbf{Avocado Email (Email)} \cite{oard2015avocado} and \textbf{Amazon Book (Book)} \cite{ni-etal-2019-justifying}. An example from Book is given in Fig.~\ref{fig.example}.


\textbf{Baselines.} We consider the state-of-the-art baselines: (1) \textbf{GTR}~\cite{ni2021large} a widely recognized retriever that operates independently of LLMs; (2) a variant of GTR, termed \textbf{Random}, in which the order of passages retrieved by GTR is randomized; (3) Point-wise score ranking \textbf{(PSR)}~\cite{izacard2022few}. This is a variant from \cite{izacard2022few} where we substitute the decoder with our LLM and apply perplexity distillation. This approach can be regarded as utilizing a reranker as a bridge model, but lacking the capability for dynamic selection; and (4) Additionally, we include a non-retrieval baseline, \textbf{Naive}, where no retrieval augmentation in the generation process.

\vspace{-3mm}
\begin{table}[h]
\centering
\resizebox{\columnwidth}{!}{
\begin{tabular}{c|cccc}
\toprule
 & \#Training & \#Val. & \#Test & Avg. \#Tokens \\ \toprule
NQ & 79,168 & 8,757 & 3,610 & 517.82 \\
HotpotQA & 68,659 & 5,600 & 5,600 & 564.83 \\
Email & 13,305 & 764 & 1,227 & 173.85 \\
Book & 20,789 & 41,331 & 41,331 & 124.52 \\ \bottomrule
\end{tabular}
}
 \vspace{-3mm}
\caption{Statistic of the 4 datasets. ``Avg. \# tokens'' indicates the average number of tokens in the prompt (which includes the query and retrieved passages). We ensure that the length does not exceed the maximum length allowed by the LLM.}
 \vspace{-3mm}
\label{tab.dataset}
\end{table}

\subsection{LLM and Hyperparamters}
\label{sec.parameter}

We select the T5-XXL (11B) \cite{DBLP:journals/jmlr/RaffelSRLNMZLL20} model as our bridge model for most experiments. In the supervised learning stage, the T5-XXL model is fine-tuned with a base learning rate of 0.001. A linear warmup scheduler is used for the first 1,000 training steps. Additionally, the square root normalized decay of the learning rate is applied. The model is trained until its performance converges on the validation set. Decoding is performed using beam search with a beam size of 4. We use the PaLM2-S model \cite{DBLP:journals/corr/abs-2305-10403}, a new state-of-the-art LLM, as our LLM. It adopts temperature sampling as the
decoding strategy. The parameters of PaLM2-S are frozen, and we set the temperature to 0 to make the output deterministic. We use Exact-Match (EM) and BLEU metrics, depending on the dataset. It is also used as the reward to train the bridge model in the RL stage. The $K$ in ``Top-K'' in Eq.~\ref{eq.topk_retrieve} is set to 5.

\subsection{Evaluation Results and Analysis}
\label{sec.exp_result}
\begin{table}[h]
\centering
\resizebox{\columnwidth}{!}{
\begin{tabular}{c|cc|cc}
\toprule
Model  & NQ    & HotpotQA & Email & Book \\
Metric & EM    & EM       & BLEU   & BLEU  \\
 \toprule
Naive  & 33.07 & 28.01    & 5.57   & 11.5  \\
Random & 43.71 & 26.10    & 8.55   & 8.61  \\
GTR    & 43.79 & 25.80    & 9.76   & 8.75  \\
PSR    & 43.60  & 25.51    & 9.08   & 9.14  \\
BGM & \textbf{45.37} & \textbf{35.64} & \textbf{10.42} & \textbf{12.07} \\ 
\bottomrule
\end{tabular}
}
 \vspace{-3mm}
\caption{Performance of all 4 datasets.}
 \vspace{-3mm}
\label{tab.overall_results}
\end{table}

\textbf{Superiority of BGM.} Table~\ref{tab.overall_results} reports the overall performance. We can see that 

(1) \textbf{The proposed BGM outperforms all the 4 baselines in all 4 datasets.} This clearly indicates BGM is effective in adapting the retrieved passages. It's noteworthy that Random and GTR perform similarly, suggesting that ranking has less impact than selection of passages does.

(2) \textbf{Compared to the Naive approach, BGM shows significant improvement overall.} An exception is observed in the Book dataset, where the improvement of BGM is less pronounced. This suggests that 
retrieval is not always essential in this context. This also explains why the Naive approach outperforms other baselines (except BGM) on the Book dataset. It's important to note that BGM still manages to show improvement by dynamically deciding whether and how many passages to select. For example, in the Book dataset, for around half of the samples (approximately 25,000), BGM selects no passages for augmentation.

(3) \textbf{Compared to the GTR approach, BGM demonstrates substantial improvement.} Compared to HotpotQA, NQ shows smaller improvement, as most instances need only one passage, which both GTR and BGM can successfully include. Conversely, the HotpotQA shows a more substantial improvement, indicating that HotpotQA may be more sensitive to irrelevant passages.  


(4) \textbf{Compared to the PSR approach, BGM again demonstrates significant improvement.} This indicates that pure reranking alone is not sufficient for the bridge model. Selection must also be taken into account. It's notable that PSR performs similarly to GTR, further suggesting that reranking alone has limited impact.


\begin{table}[h]
\centering
\resizebox{\columnwidth}{!}{
\begin{tabular}{c|cc|cc}
\toprule
Model  & NQ    & HotpotQA & Email & Book \\
Metric & EM    & EM       & BLEU   & BLEU  \\
 \toprule
Naive & 33.07 & 28.01 & 5.57 & 11.5 \\
PSR & 43.60 & 25.51 & 9.08 & 9.14 \\
PSR (Top1) & 42.02 & 32.69 & 7.28 & 11.53 \\
PSR (Top2) & 42.54 & 31.05 & 7.77 & 10.11 \\
PSR (Top3) & 42.85 & 32.71 & 8.21 & 9.70 \\
PSR (Top4) & 43.71 & 32.37 & 8.26 & 9.11 \\
BGM & \textbf{45.37} & \textbf{35.64} & \textbf{10.42} & \textbf{12.07} \\ 
\bottomrule
\end{tabular}
}
 \vspace{-3mm}
\caption{Ablation - threshold the PSR.
}
 \vspace{-7mm}
\label{tab.abalation_PSR}
\end{table}

\begin{table}[h]
\centering
\resizebox{\columnwidth}{!}{
\begin{tabular}{c|cc|cc}
\toprule
Silver Data & NQ & HotpotQA & Email & Book \\
Metric & EM & EM & BLEU & BLEU \\
 \toprule
GTR & 43.79 & 25.80 & 9.76 & 8.75 \\
PSR & 43.68 & 29.73 & 10.1 & 10.35 \\
Greedy (BGM) & \textbf{45.37} & \textbf{35.64} & \textbf{10.42} & \textbf{12.07} \\
\bottomrule
\end{tabular}
}
 \vspace{-3mm}
\caption{Ablation - different SPS for SL.
}
 \vspace{-7mm}
\label{tab.abalation_SL_objective}
\end{table}

\begin{table}[h]
\centering
\resizebox{\columnwidth}{!}{
\begin{tabular}{c|cc|cc}
\toprule
Model & NQ & HotpotQA & Email & Book \\
Metric & EM & EM & BLEU & BLEU \\
 \toprule
GTR & 43.79 & 25.8 & 9.76 & 8.75 \\
BGM (SL only) & 39.44 & 34.26 & 8.62 & 12.05 \\
BGM & \textbf{45.37} & \textbf{35.64} & \textbf{10.42} & \textbf{12.07} \\
\bottomrule
\end{tabular}
}
 \vspace{-3mm}
\caption{Ablation - SL only.
}
\label{tab.abalation_SL_only}
\end{table}

\begin{table}[h]
\centering
\resizebox{\columnwidth}{!}{
\begin{tabular}{c|cc|cc}
\toprule
Model & NQ & HotpotQA & Email & Book \\
Metric & EM & EM & BLEU & BLEU \\
 \toprule
GTR & 43.79 & 25.8 & 9.76 & 8.75 \\
FLAN-T5-Large & 44.15 & 35.87 & 10.18 & 10.19 \\
FLAN-T5-XL & 44.87 & 35.41 & 9.64 & 10.7 \\
FLAN-T5-XXL & \textbf{45.37} & \textbf{35.64} & \textbf{10.42} & \textbf{12.07} \\
\bottomrule
\end{tabular}
}
 \vspace{-3mm}
\caption{Ablation - different BGM bridge model size.
}
 \vspace{-4mm}
\label{tab.abalation_bridge model_size}
\end{table}

\begin{table}[h]
\centering
\resizebox{\columnwidth}{!}{
\begin{tabular}{c|cc|cc}
\toprule
Model & \multicolumn{2}{c|}{Palm2-XXS}& \multicolumn{2}{c}{Palm2-S} \\
Data & NQ & HotpotQA & NQ & HotpotQA \\
Metric & EM & EM & EM & EM \\
\toprule
Naive & 12.13 & 14.57 & 33.07 & 28.01 \\
Random & 31.19 & 24.41 & 43.71 & 26.10 \\
GTR & 31.91 & 23.17 & 43.79 & 25.80 \\
PSR & 32.04 & 22.53 & 43.60  & 25.51 \\
BGM & \textbf{39.88} & \textbf{28.69} & \textbf{45.37} & \textbf{35.64} \\
\bottomrule
\end{tabular}
}
 \vspace{-3mm}
\caption{Ablation - different LLM size.
}
 \vspace{-5mm}
\label{tab.abalation_llm_size}
\end{table}

\begin{table}[h]
\centering
\resizebox{\columnwidth}{!}{
\begin{tabular}{c|cc|cc}
\toprule
Model & NQ & HotpotQA & Email & Book \\
Metric & EM & EM & BLEU & BLEU \\
\hline
\multicolumn{5}{c}{\textbf{Test on Palm2-S}} \\
\hline
BGM (in-domain) & \textbf{45.37} & \textbf{35.64} & \textbf{10.42} & \textbf{12.07} \\
\begin{tabular}[c]{@{}c@{}}BGM \\ (Trained on NQ)\end{tabular} & --- & 33.42 & 5.66 & 11.22 \\
\begin{tabular}[c]{@{}c@{}}BGM \\ (Trained on Email)\end{tabular} & 35.59 & 27.98 & --- & 11.38 \\
\hline
\multicolumn{5}{c}{\textbf{Test on Palm2-XXS}} \\
\hline
\begin{tabular}[c]{@{}c@{}}BGM \\ (Trained with Palm2-XXS)\end{tabular} &  \textbf{39.88} & \textbf{28.69} & --- & --- \\
\begin{tabular}[c]{@{}c@{}}BGM \\ (Trained with Palm2-S)\end{tabular} & 30.63 & 24.55 & --- & --- \\
\bottomrule
\end{tabular}
}
 \vspace{-3mm}
\caption{Ablation - the generability of BGM across various datasets and LLM sizes.
}
 \vspace{-5mm}
\label{tab.abalation_generalizable}
\end{table}

\textbf{Understanding BGM.} Next, we study a few research questions about BGM and RAG.



(1) \textbf{Can we perform effectively passage selection for RAG by simply thresholding PSR?} We conducted an ablation experiment using various thresholds for PSR, as shown in Table~\ref{tab.abalation_PSR}. ``Top-K'' in rows 3 to 6 denotes selecting only the top-k passages from PSR reranked passages. The results vary but are consistently lower than BGM's. This suggests that a naive manual threshold applied to the reranking model is insufficient to achieve the desired objectives. Therefore, it is necessary to consider both reranking and dynamic selection simultaneously.



(2) \textbf{How does different SPS (sliver passage sequence) used for SL affect BGM performance?} In Table~\ref{tab.overall_results}, we employed greedy search passage sequences as SPS for SL (Sec.~\ref{sec.sl}). Here, we explore whether this approach is superior to others. Our ablation experiment in Table~\ref{tab.abalation_SL_objective} involved using various types of SPS. The first row, labeled ``GTR'', indicates the use of GTR's retrieved passages as SPS. Similarly, ``PSR'' refers to using PSR reranked passages as SPS. In the final row, we used greedy search passage sequences, representing the proposed version of BGM. The results demonstrate that the quality of SPS significantly affects downstream task performance. While using PSR as SPS shows improvement over GTR on three datasets, Greedy (BGM) further enhances performance and achieves the best results. Identifying potentially better SPS is left for future work.


(3) \textbf{How helpful is RL to BGM?} BGM integrates SL and RL, and we aim to assess the effectiveness of each component. An ablation experiment is detailed in Table~\ref{tab.abalation_SL_only}. We can observe that BGM, when operating with only SL, performs significantly worse than the full BGM model in NQ, HotpotQA, and Email, and in some cases, it even underperforms GTR. This suggests that SL alone is inadequate, highlighting the necessity of incorporating RL. Note that removing SL is likely ineffective, leading to a poorly initialized policy model for RL with an excessively large search space.


(4) \textbf{How does the size of bridge model affect the final performance?} In Table~\ref{tab.overall_results}, we utilized Flan-T5-XXL (11B) as the bridge model, which is already smaller than the PaLM2-S. However, we are curious about the feasibility of using an even more lightweight LM as the bridge model. The ablation study shown in Table~\ref{tab.abalation_bridge model_size} presents the outcomes of employing bridge models of various sizes. It is evident that all three sizes (large, XL, and XXL) surpass the performance of the setup without a bridge model (i.e., GTR), with the largest size yielding the best results. This demonstrates that a bridge model is beneficial even at a smaller scale, and that larger sizes lead to further improvements.

(5) \textbf{How does the size of LLMs affect the final performance?} In Table~\ref{tab.overall_results}, Palm2-S was used as the LLM. We are interested in evaluating the effectiveness of the bridge model with different sizes of LLMs. We conducted experiments using Palm2-XXS in Table~\ref{tab.abalation_llm_size}. It is important to note that the smaller LLM struggles with personalized generation datasets (i.e., Email and Book), resulting in BLEU scores lower than 1\%. We opted not to report these results as they may not accurately reflect the trend. However, observations from NQ and HotpotQA suggest that BGM significantly outperforms all baselines by a large margin. This indicates that BGM is effective even with a smaller LLM.

(6) \textbf{Can a bridge model generalize to different datasets and LLMs?} Our experiments above demonstrate the effectiveness of the bridge model when trained ``in-domain'' (i.e., trained and tested on the same dataset) using PaLM2-S as the LLM. 
A more ambitious goal is to extend this performance to new datasets or LLMs without extra training. We conducted ablation experiments to investigate bridge model generalizability: Table~\ref{tab.abalation_generalizable}'s upper section shows the results across different datasets. Row 2 shows the performance of the bridge model when trained exclusively on the NQ dataset and then tested on three other unseen datasets. Similarly, row 3 shows training BGM solely on the Email dataset and testing it on the other three. In all cases, performance falls short of BGM's when trained and tested on the same datasets. This is expected, given the lack of techniques for dataset generalization, a topic we leave for future work.
 
In the lower section of Table~\ref{tab.abalation_generalizable}, we present the results of BGM when tested on Palm2-XXS, but trained on Palm2-S. Comparing with results of BGM both trained and tested on Palm2-XXS (see row 1 of the bottom section), the mismatch between the training and testing LLMs leads to a significant decline in performance. This suggests that BGM's ability to generalize across different LLMs is currently limited. Addressing this is considered an important direction for future research.

(7) \textbf{Case Studies.} We provided examples of GTR, PSR, and BGM in Table~\ref{ap_tab.case} in Appendix for the NQ dataset. For \textbf{question I}, both GTR and PSR yield the same incorrect output, even though both include a relevant passage for RAG. Only BGM provides the correct answer, indicating that additional irrelevant context can be noisy and detrimental to RAG's performance.
In \textbf{question II}, none of the candidate passages contain the answer (they discuss FaZe Clan and the number of subscribers, but do not identify who has the most subscribers). GTR and PSR provide incorrect answers, as their additional context is unhelpful. In contrast, BGM opts not to select any passages and answers the question using its own memory, resulting in the correct answer. This demonstrates that retrieval-augmented processes are not always necessary, and BGM is capable of handling such cases.

\section{Conclusion}

This paper demonstrates the need to bridge the preference gap between retrievers and LLMs, which has various levels of impact on ranking and selection in RAG systems. We propose a bridge model, BGM, to adapt the output of a frozen retriever for frozen LLMs, formatting the task as a seq2seq problem. BGM chains supervised and reinforcement learning, for dense supervisions and end-to-end training. Our extensive experiments validate BGM's effectiveness.


\section{Limitations}
While effective, BGM has certain potential limitations. Firstly, BGM is limited in terms of generalization across datasets and domains, as shown in Sec.~\ref{sec.exp_result}. Secondly, the silver passage sequences for supervised learning is synthesized using greedy search. Although Sec.~\ref{sec.exp_result} demonstrates its effectiveness, it is worth exploring other synthesizing approaches. We leave these explorations to the future work, as bridging the preference gap is an critical but new problem in RAG.

\section{Ethics Statement}
This paper proposes a novel bridge model that can effectively bridge the preference gap between retrievers and Large Language Models (LLMs). We do not anticipate any negative consequences for individuals as a result of this research. In the event of system failure, the primary outcome would be suboptimal performance of an LLM, and we do not foresee notable ethical implications.

\bibliography{anthology,custom}

\begin{thebibliography}{33}
\expandafter\ifx\csname natexlab\endcsname\relax\def\natexlab#1{#1}\fi

\bibitem[{Adolphs et~al.(2021)Adolphs, Boerschinger, Buck, Huebscher, Ciaramita, Espeholt, Hofmann, Kilcher, Rothe, Sessa et~al.}]{adolphs2021boosting}
Leonard Adolphs, Benjamin Boerschinger, Christian Buck, Michelle~Chen Huebscher, Massimiliano Ciaramita, Lasse Espeholt, Thomas Hofmann, Yannic Kilcher, Sascha Rothe, Pier~Giuseppe Sessa, et~al. 2021.
\newblock Boosting search engines with interactive agents.
\newblock \emph{arXiv preprint arXiv:2109.00527}.

\bibitem[{Anil et~al.(2023)Anil, Dai, Firat, Johnson, Lepikhin, Passos, Shakeri, Taropa, Bailey, Chen, Chu, Clark, Shafey, Huang, Meier{-}Hellstern, Mishra, Moreira, Omernick, Robinson, Ruder, Tay, Xiao, Xu, Zhang, {\'{A}}brego, Ahn, Austin, Barham, Botha, Bradbury, Brahma, Brooks, Catasta, Cheng, Cherry, Choquette{-}Choo, Chowdhery, Crepy, Dave, Dehghani, Dev, Devlin, D{\'{\i}}az, Du, Dyer, Feinberg, Feng, Fienber, Freitag, Garcia, Gehrmann, Gonzalez, and et~al.}]{DBLP:journals/corr/abs-2305-10403}
Rohan Anil, Andrew~M. Dai, Orhan Firat, Melvin Johnson, Dmitry Lepikhin, Alexandre Passos, Siamak Shakeri, Emanuel Taropa, Paige Bailey, Zhifeng Chen, Eric Chu, Jonathan~H. Clark, Laurent~El Shafey, Yanping Huang, Kathy Meier{-}Hellstern, Gaurav Mishra, Erica Moreira, Mark Omernick, Kevin Robinson, Sebastian Ruder, Yi~Tay, Kefan Xiao, Yuanzhong Xu, Yujing Zhang, Gustavo~Hern{\'{a}}ndez {\'{A}}brego, Junwhan Ahn, Jacob Austin, Paul Barham, Jan~A. Botha, James Bradbury, Siddhartha Brahma, Kevin Brooks, Michele Catasta, Yong Cheng, Colin Cherry, Christopher~A. Choquette{-}Choo, Aakanksha Chowdhery, Cl{\'{e}}ment Crepy, Shachi Dave, Mostafa Dehghani, Sunipa Dev, Jacob Devlin, Mark D{\'{\i}}az, Nan Du, Ethan Dyer, Vladimir Feinberg, Fangxiaoyu Feng, Vlad Fienber, Markus Freitag, Xavier Garcia, Sebastian Gehrmann, Lucas Gonzalez, and et~al. 2023.
\newblock Palm 2 technical report.
\newblock \emph{CoRR}, abs/2305.10403.

\bibitem[{Bacciu et~al.(2023)Bacciu, Cocunasu, Siciliano, Silvestri, Tonellotto, and Trappolini}]{bacciu2023rraml}
Andrea Bacciu, Florin Cocunasu, Federico Siciliano, Fabrizio Silvestri, Nicola Tonellotto, and Giovanni Trappolini. 2023.
\newblock Rraml: Reinforced retrieval augmented machine learning.
\newblock \emph{arXiv preprint arXiv:2307.12798}.

\bibitem[{Borgeaud et~al.(2022)Borgeaud, Mensch, Hoffmann, Cai, Rutherford, Millican, Van Den~Driessche, Lespiau, Damoc, Clark et~al.}]{borgeaud2022improving}
Sebastian Borgeaud, Arthur Mensch, Jordan Hoffmann, Trevor Cai, Eliza Rutherford, Katie Millican, George~Bm Van Den~Driessche, Jean-Baptiste Lespiau, Bogdan Damoc, Aidan Clark, et~al. 2022.
\newblock Improving language models by retrieving from trillions of tokens.
\newblock In \emph{International conference on machine learning}, pages 2206--2240. PMLR.

\bibitem[{De~Jong et~al.(2023)De~Jong, Zemlyanskiy, FitzGerald, Ainslie, Sanghai, Sha, and Cohen}]{de2023pre}
Michiel De~Jong, Yury Zemlyanskiy, Nicholas FitzGerald, Joshua Ainslie, Sumit Sanghai, Fei Sha, and William~W Cohen. 2023.
\newblock Pre-computed memory or on-the-fly encoding? a hybrid approach to retrieval augmentation makes the most of your compute.
\newblock In \emph{International Conference on Machine Learning}, pages 7329--7342. PMLR.

\bibitem[{de~Jong et~al.(2023)de~Jong, Zemlyanskiy, FitzGerald, Sanghai, Cohen, and Ainslie}]{de2023glimmer}
Michiel de~Jong, Yury Zemlyanskiy, Nicholas FitzGerald, Sumit Sanghai, William~W Cohen, and Joshua Ainslie. 2023.
\newblock Glimmer: generalized late-interaction memory reranker.
\newblock \emph{arXiv preprint arXiv:2306.10231}.

\bibitem[{Guu et~al.(2020)Guu, Lee, Tung, Pasupat, and Chang}]{guu2020retrieval}
Kelvin Guu, Kenton Lee, Zora Tung, Panupong Pasupat, and Mingwei Chang. 2020.
\newblock Retrieval augmented language model pre-training.
\newblock In \emph{International conference on machine learning}, pages 3929--3938. PMLR.

\bibitem[{Izacard et~al.(2021)Izacard, Caron, Hosseini, Riedel, Bojanowski, Joulin, and Grave}]{izacard2021unsupervised}
Gautier Izacard, Mathilde Caron, Lucas Hosseini, Sebastian Riedel, Piotr Bojanowski, Armand Joulin, and Edouard Grave. 2021.
\newblock Unsupervised dense information retrieval with contrastive learning.
\newblock \emph{arXiv preprint arXiv:2112.09118}.

\bibitem[{Izacard and Grave(2020)}]{izacard2020leveraging}
Gautier Izacard and Edouard Grave. 2020.
\newblock Leveraging passage retrieval with generative models for open domain question answering.
\newblock \emph{arXiv preprint arXiv:2007.01282}.

\bibitem[{Izacard et~al.(2022)Izacard, Lewis, Lomeli, Hosseini, Petroni, Schick, Dwivedi-Yu, Joulin, Riedel, and Grave}]{izacard2022few}
Gautier Izacard, Patrick Lewis, Maria Lomeli, Lucas Hosseini, Fabio Petroni, Timo Schick, Jane Dwivedi-Yu, Armand Joulin, Sebastian Riedel, and Edouard Grave. 2022.
\newblock Few-shot learning with retrieval augmented language models.
\newblock \emph{arXiv preprint arXiv:2208.03299}.

\bibitem[{Karpukhin et~al.(2020)Karpukhin, O{\u{g}}uz, Min, Lewis, Wu, Edunov, Chen, and Yih}]{karpukhin2020dense}
Vladimir Karpukhin, Barlas O{\u{g}}uz, Sewon Min, Patrick Lewis, Ledell Wu, Sergey Edunov, Danqi Chen, and Wen-tau Yih. 2020.
\newblock Dense passage retrieval for open-domain question answering.
\newblock \emph{arXiv preprint arXiv:2004.04906}.

\bibitem[{Khandelwal et~al.(2020)Khandelwal, Levy, Jurafsky, Zettlemoyer, and Lewis}]{Khandelwal2020Generalization}
Urvashi Khandelwal, Omer Levy, Dan Jurafsky, Luke Zettlemoyer, and Mike Lewis. 2020.
\newblock \href {https://openreview.net/forum?id=HklBjCEKvH} {Generalization through memorization: Nearest neighbor language models}.
\newblock In \emph{International Conference on Learning Representations}.

\bibitem[{Kwiatkowski et~al.(2019)Kwiatkowski, Palomaki, Redfield, Collins, Parikh, Alberti, Epstein, Polosukhin, Devlin, Lee, Toutanova, Jones, Kelcey, Chang, Dai, Uszkoreit, Le, and Petrov}]{kwiatkowski-etal-2019-natural}
Tom Kwiatkowski, Jennimaria Palomaki, Olivia Redfield, Michael Collins, Ankur Parikh, Chris Alberti, Danielle Epstein, Illia Polosukhin, Jacob Devlin, Kenton Lee, Kristina Toutanova, Llion Jones, Matthew Kelcey, Ming-Wei Chang, Andrew~M. Dai, Jakob Uszkoreit, Quoc Le, and Slav Petrov. 2019.
\newblock \href {https://doi.org/10.1162/tacl_a_00276} {Natural questions: A benchmark for question answering research}.
\newblock \emph{Transactions of the Association for Computational Linguistics}, 7:452--466.

\bibitem[{Lewis et~al.(2020)Lewis, Perez, Piktus, Petroni, Karpukhin, Goyal, K{\"u}ttler, Lewis, Yih, Rockt{\"a}schel et~al.}]{lewis2020retrieval}
Patrick Lewis, Ethan Perez, Aleksandra Piktus, Fabio Petroni, Vladimir Karpukhin, Naman Goyal, Heinrich K{\"u}ttler, Mike Lewis, Wen-tau Yih, Tim Rockt{\"a}schel, et~al. 2020.
\newblock Retrieval-augmented generation for knowledge-intensive nlp tasks.
\newblock \emph{Advances in Neural Information Processing Systems}, 33:9459--9474.

\bibitem[{Li et~al.(2023)Li, Zhang, Mei, Wang, Hombaiah, Liang, and Bendersky}]{li2023teach}
Cheng Li, Mingyang Zhang, Qiaozhu Mei, Yaqing Wang, Spurthi~Amba Hombaiah, Yi~Liang, and Michael Bendersky. 2023.
\newblock Teach llms to personalize--an approach inspired by writing education.
\newblock \emph{arXiv preprint arXiv:2308.07968}.

\bibitem[{Liu et~al.(2023)Liu, Lin, Hewitt, Paranjape, Bevilacqua, Petroni, and Liang}]{liu2023lost}
Nelson~F Liu, Kevin Lin, John Hewitt, Ashwin Paranjape, Michele Bevilacqua, Fabio Petroni, and Percy Liang. 2023.
\newblock Lost in the middle: How language models use long contexts.
\newblock \emph{arXiv preprint arXiv:2307.03172}.

\bibitem[{Ni et~al.(2019)Ni, Li, and McAuley}]{ni-etal-2019-justifying}
Jianmo Ni, Jiacheng Li, and Julian McAuley. 2019.
\newblock \href {https://doi.org/10.18653/v1/D19-1018} {Justifying recommendations using distantly-labeled reviews and fine-grained aspects}.
\newblock In \emph{Proceedings of the 2019 Conference on Empirical Methods in Natural Language Processing and the 9th International Joint Conference on Natural Language Processing (EMNLP-IJCNLP)}, pages 188--197, Hong Kong, China. Association for Computational Linguistics.

\bibitem[{Ni et~al.(2021)Ni, Qu, Lu, Dai, {\'A}brego, Ma, Zhao, Luan, Hall, Chang et~al.}]{ni2021large}
Jianmo Ni, Chen Qu, Jing Lu, Zhuyun Dai, Gustavo~Hern{\'a}ndez {\'A}brego, Ji~Ma, Vincent~Y Zhao, Yi~Luan, Keith~B Hall, Ming-Wei Chang, et~al. 2021.
\newblock Large dual encoders are generalizable retrievers.
\newblock \emph{arXiv preprint arXiv:2112.07899}.

\bibitem[{Nogueira and Cho(2017)}]{nogueira2017task}
Rodrigo Nogueira and Kyunghyun Cho. 2017.
\newblock Task-oriented query reformulation with reinforcement learning.
\newblock \emph{arXiv preprint arXiv:1704.04572}.

\bibitem[{Oard et~al.(2015)Oard, Webber, Kirsch, and Golitsynskiy}]{oard2015avocado}
Douglas Oard, William Webber, David Kirsch, and Sergey Golitsynskiy. 2015.
\newblock Avocado research email collection.
\newblock \emph{Philadelphia: Linguistic Data Consortium}.

\bibitem[{Raffel et~al.(2020{\natexlab{a}})Raffel, Shazeer, Roberts, Lee, Narang, Matena, Zhou, Li, and Liu}]{raffel2020exploring}
Colin Raffel, Noam Shazeer, Adam Roberts, Katherine Lee, Sharan Narang, Michael Matena, Yanqi Zhou, Wei Li, and Peter~J Liu. 2020{\natexlab{a}}.
\newblock Exploring the limits of transfer learning with a unified text-to-text transformer.
\newblock \emph{The Journal of Machine Learning Research}, 21(1):5485--5551.

\bibitem[{Raffel et~al.(2020{\natexlab{b}})Raffel, Shazeer, Roberts, Lee, Narang, Matena, Zhou, Li, and Liu}]{DBLP:journals/jmlr/RaffelSRLNMZLL20}
Colin Raffel, Noam Shazeer, Adam Roberts, Katherine Lee, Sharan Narang, Michael Matena, Yanqi Zhou, Wei Li, and Peter~J. Liu. 2020{\natexlab{b}}.
\newblock Exploring the limits of transfer learning with a unified text-to-text transformer.
\newblock \emph{J. Mach. Learn. Res.}

\bibitem[{Shi et~al.(2023{\natexlab{a}})Shi, Chen, Misra, Scales, Dohan, Chi, Sch{\"a}rli, and Zhou}]{shi2023large}
Freda Shi, Xinyun Chen, Kanishka Misra, Nathan Scales, David Dohan, Ed~H Chi, Nathanael Sch{\"a}rli, and Denny Zhou. 2023{\natexlab{a}}.
\newblock Large language models can be easily distracted by irrelevant context.
\newblock In \emph{International Conference on Machine Learning}, pages 31210--31227. PMLR.

\bibitem[{Shi et~al.(2023{\natexlab{b}})Shi, Min, Yasunaga, Seo, James, Lewis, Zettlemoyer, and Yih}]{shi2023replug}
Weijia Shi, Sewon Min, Michihiro Yasunaga, Minjoon Seo, Rich James, Mike Lewis, Luke Zettlemoyer, and Wen-tau Yih. 2023{\natexlab{b}}.
\newblock Replug: Retrieval-augmented black-box language models.
\newblock \emph{arXiv preprint arXiv:2301.12652}.

\bibitem[{Wei et~al.(2017)Wei, Xu, Lan, Guo, and Cheng}]{wei2017reinforcement}
Zeng Wei, Jun Xu, Yanyan Lan, Jiafeng Guo, and Xueqi Cheng. 2017.
\newblock Reinforcement learning to rank with markov decision process.
\newblock In \emph{Proceedings of the 40th international ACM SIGIR conference on research and development in information retrieval}.

\bibitem[{Wu et~al.(2022)Wu, Rabe, Hutchins, and Szegedy}]{wu2022memorizing}
Yuhuai Wu, Markus~N Rabe, DeLesley Hutchins, and Christian Szegedy. 2022.
\newblock Memorizing transformers.
\newblock \emph{arXiv preprint arXiv:2203.08913}.

\bibitem[{Wu et~al.(2021)Wu, Luan, Rashkin, Reitter, Hajishirzi, Ostendorf, and Tomar}]{wu2021conqrr}
Zeqiu Wu, Yi~Luan, Hannah Rashkin, David Reitter, Hannaneh Hajishirzi, Mari Ostendorf, and Gaurav~Singh Tomar. 2021.
\newblock Conqrr: Conversational query rewriting for retrieval with reinforcement learning.
\newblock \emph{arXiv preprint arXiv:2112.08558}.

\bibitem[{Xia et~al.(2017)Xia, Xu, Lan, Guo, Zeng, and Cheng}]{xia2017adapting}
Long Xia, Jun Xu, Yanyan Lan, Jiafeng Guo, Wei Zeng, and Xueqi Cheng. 2017.
\newblock Adapting markov decision process for search result diversification.
\newblock In \emph{Proceedings of the 40th international ACM SIGIR conference on research and development in information retrieval}.

\bibitem[{Xu et~al.(2023)Xu, Shi, and Choi}]{xu2023recomp}
Fangyuan Xu, Weijia Shi, and Eunsol Choi. 2023.
\newblock Recomp: Improving retrieval-augmented lms with compression and selective augmentation.
\newblock \emph{arXiv preprint arXiv:2310.04408}.

\bibitem[{Xu et~al.(2020)Xu, Wei, Xia, Lan, Yin, Cheng, and Wen}]{xu2020reinforcement}
Jun Xu, Zeng Wei, Long Xia, Yanyan Lan, Dawei Yin, Xueqi Cheng, and Ji-Rong Wen. 2020.
\newblock Reinforcement learning to rank with pairwise policy gradient.
\newblock In \emph{Proceedings of the 43rd International ACM SIGIR Conference on Research and Development in Information Retrieval}, pages 509--518.

\bibitem[{Yang et~al.(2018)Yang, Qi, Zhang, Bengio, Cohen, Salakhutdinov, and Manning}]{DBLP:conf/emnlp/Yang0ZBCSM18}
Zhilin Yang, Peng Qi, Saizheng Zhang, Yoshua Bengio, William~W. Cohen, Ruslan Salakhutdinov, and Christopher~D. Manning. 2018.
\newblock Hotpotqa: {A} dataset for diverse, explainable multi-hop question answering.
\newblock In \emph{EMNLP}.

\bibitem[{Yasunaga et~al.(2023)Yasunaga, Aghajanyan, Shi, James, Leskovec, Liang, Lewis, Zettlemoyer, and Yih}]{yasunaga2023retrieval}
Michihiro Yasunaga, Armen Aghajanyan, Weijia Shi, Richard James, Jure Leskovec, Percy Liang, Mike Lewis, Luke Zettlemoyer, and Wen-tau Yih. 2023.
\newblock Retrieval-augmented multimodal language modeling.

\bibitem[{Zeng et~al.(2018)Zeng, Xu, Lan, Guo, and Cheng}]{zeng2018multi}
Wei Zeng, Jun Xu, Yanyan Lan, Jiafeng Guo, and Xueqi Cheng. 2018.
\newblock Multi page search with reinforcement learning to rank.
\newblock In \emph{Proceedings of the 2018 ACM SIGIR international conference on theory of information retrieval}, pages 175--178.

\end{thebibliography}
\bibliographystyle{acl_natbib}

\appendix

\section{Case Studies}
\small
\onecolumn
\begin{longtable}{p{\textwidth}}
\specialrule{.2em}{.1em}{.1em}
\textbf{Question I} \\
\hline
Where does new crust come from in sea floor spreading \\
\hline
\textbf{Answer} \\
\hline
``basaltic magma'' or ``volcanic activity'' \\
\hline\hline
\textbf{GTR Context} \\
\hline
\textbf{Given the passage titles and contexts below:} \\ 
\textbf{Title:} Seafloor spreading;\\ 
\textbf{Context:} Seafloor spreading Seafloor spreading is a process that occurs at mid-ocean ridges, where new oceanic crust is formed through volcanic activity and then gradually moves away from the ridge. Earlier theories (e.g. by Alfred Wegener and Alexander du Toit) of continental drift postulated that continents "ploughed" through the sea. The idea that the seafloor itself moves (and also carries the continents with it) 
as it expands from a central axis was proposed by Harry Hess from Princeton University in the 1960s. The theory is well accepted now, and the phenomenon is known to be caused by convection currents in the; \\
\textbf{Title:} Oceanic crust; \\
\textbf{Context:} frozen in the basalt. A symmetrical pattern of positive and negative magnetic lines emanates from the mid-ocean ridge. New rock is formed by magma at the mid-ocean ridges, and the ocean floor spreads out from this point. When the magma cools to form rock, its magnetic polarity is aligned with the then-current positions of the magnetic poles of the Earth. New magma then forces the older cooled magma away from the ridge. This process results in parallel sections of oceanic crust of alternating magnetic polarity. Oceanic crust Oceanic crust is the uppermost layer of the oceanic portion of a tectonic; \\
\textbf{Title:} Seafloor spreading; \\
\textbf{Context:} $\sim$Tx/L, where L is the distance between the ridge to the continental shelf (roughly half the ocean width), and T is the ocean age. Seafloor spreading Seafloor spreading is a process that occurs at mid-ocean ridges, where new oceanic crust is formed through volcanic activity and then gradually moves away from the ridge. Earlier theories (e.g. by Alfred Wegener and Alexander du Toit) of continental drift postulated that continents "ploughed" through the sea. The idea that the seafloor itself moves (and also carries the continents with it) as it expands from a central axis was proposed by Harry Hess; \\
\textbf{Title:} Seafloor spreading; \\
\textbf{Context:} As new seafloor forms and spreads apart from the mid-ocean ridge it slowly cools over time. Older seafloor is therefore colder than new seafloor, and older oceanic basins deeper than new oceanic basins due to isostasy. If the diameter of the earth remains relatively constant despite the production of new crust, a mechanism must exist by which crust is also destroyed. The destruction of oceanic crust occurs at subduction zones where oceanic crust is  forced under either continental crust or oceanic crust. Today, the Atlantic basin is actively spreading at the Mid-Atlantic Ridge. Only a small portion of the oceanic; \\
\textbf{Title:} Seafloor spreading; \\
\textbf{Context:} other areas), material from the upper mantle rises through the faults  between oceanic plates to form new crust as the plates move away from each other, a phenomenon first observed as continental drift. When Alfred Wegener first presented a hypothesis of continental drift in 1912, he suggested that continents ploughed through the ocean crust. This was impossible: oceanic crust is both more dense and more rigid than continental crust. Accordingly, Wegener's theory wasn't taken very seriously, especially in the United States. Since then, it has been shown that the motion of the continents is linked to seafloor spreading by the; \\
\textbf{What is the answer of the following question: }\\
where does new crust come from in sea floor spreading \\
\hline
\textbf{GTR Prediction}: the upper mantle \\
\hline
\hline
\textbf{PSR Context} \\
\hline
\textbf{Given the passage titles and contexts below:} \\ \textbf{Title:} Seafloor spreading; \\
\textbf{Context:} Seafloor spreading Seafloor spreading is a process that occurs at mid-ocean ridges, where new oceanic crust is formed through volcanic activity and then gradually moves away from the ridge.  Earlier theories (e.g. by Alfred Wegener and Alexander du Toit) of continental drift postulated that continents "ploughed" through the sea.  The idea that the seafloor itself moves (and also carries the continents with it) as it expands from a central axis was proposed by Harry Hess from Princeton University in the 1960s. The theory is well accepted now, and the phenomenon is known to be caused by convection currents in the; 
\textbf{Title:} Seafloor spreading;\\
\textbf{Context:} $\sim$Tx/L, where L is the distance between the ridge to the continental shelf (roughly half the ocean width), and T is the ocean age. Seafloor spreading Seafloor spreading is a process that occurs at mid-ocean ridges, where new oceanic crust is formed through volcanic activity and then gradually moves away from the ridge. Earlier theories (e.g. by Alfred Wegener and Alexander du Toit) of continental drift postulated that continents "ploughed" through the sea. The idea that the seafloor itself moves (and also carries the continents with it) as it expands from a central axis was proposed by Harry Hess;
\textbf{Title:} Seafloor spreading;\\
\textbf{Context:} other areas), material from the upper mantle rises through the faults between oceanic plates to form new crust as the plates move away from each other, a phenomenon first observed as continental drift. When Alfred Wegener first presented a hypothesis of continental drift in 1912, he suggested that continents ploughed through the ocean crust. This was impossible: oceanic crust is both more dense and more rigid than continental crust.  Accordingly, Wegener's theory wasn't taken very seriously, especially in the United States. Since then, it has been shown that the motion of the continents is linked to seafloor spreading by the; \\
\textbf{Title:} Oceanic crust;\\
\textbf{Context:} frozen in the basalt. A symmetrical pattern of positive and negative magnetic lines emanates from the mid-ocean ridge. New rock is formed by magma at the mid-ocean ridges, and the ocean floor spreads out from this point. When the magma cools to form rock, its magnetic polarity is aligned with the then-current positions of the magnetic poles of the Earth. New magma then forces the older cooled magma away from the ridge. This process results in parallel sections of oceanic crust of alternating magnetic polarity. Oceanic crust Oceanic crust is the uppermost layer of the oceanic portion of a tectonic; \\
\textbf{Title:} Seafloor spreading; \\
\textbf{Context:} As new seafloor forms and spreads apart from the mid-ocean ridge it slowly cools over time. Older seafloor is therefore colder than new seafloor, and older oceanic basins deeper than new oceanic basins due to isostasy. If the diameter of the earth remains relatively constant despite the production of new crust, a mechanism must exist by which crust is also destroyed. The destruction of oceanic crust occurs at subduction zones where oceanic crust is forced under either continental crust or oceanic crust. Today, the Atlantic basin is actively spreading at the Mid-Atlantic Ridge. Only a small portion of the oceanic; \\ 
\textbf{What is the answer of the following question:} \\ 
where does new crust come from in sea floor spreading \\
\hline
\textbf{PSR Prediction}: the upper mantle \\
\hline
\hline
\textbf{BGM Context} \\
\hline
\textbf{Given the passage titles and contexts below:} \\ 
\textbf{Title:} Seafloor spreading;\\ 
\textbf{Context:} Seafloor spreading Seafloor spreading is a process that occurs at mid-ocean ridges, where new oceanic crust is formed through volcanic activity and then gradually moves away from the ridge. Earlier theories (e.g. by Alfred Wegener and Alexander du Toit) of continental drift postulated that continents "ploughed" through the sea. The idea that the seafloor itself moves (and also carries the continents with it) as it expands from a central axis was proposed by Harry Hess from Princeton University in the 1960s. The theory is well accepted now, and the phenomenon is known to be caused by convection currents in the; \\
\textbf{What is the answer of the following question:} \\ 
where does new crust come from in sea floor spreading \\
\hline
\textbf{BGM Prediction}: volcanic activity \\
\specialrule{.2em}{.1em}{.1em}
\textbf{Question II} \\
\hline
who has the most subscribers in faze clan\\
\hline
\textbf{Answer} \\
\hline
``FaZe Rug'' \\
\hline
\hline
\textbf{GTR Context} \\
\hline
\textbf{Given the passage titles and contexts below:}\\
\textbf{Title:} FaZe Clan;\\ \textbf{Context:}  "FaZe 2.0" by FaZe members and fans. The FaZe "CS:GO" team went on to become one of the most successful rosters for the 2017/2018 seasons. FaZe Clan is the most popular esports organization in the world, based on the organization's social media following. As of November 28, 2018, FaZe Clan and its members together have 82 million YouTube subscribers, 11.2 billion YouTube views, 11.3 million Twitch followers, 130 million Twitch views, 43.1 million Twitter followers, 45.8 million Instagram followers, 2.8 million Facebook likes and followers. FaZe Clan has made \$6,148,290.91 from esport tournament prize pools alone. FaZe Clan started on;\\ \textbf{Title:}  FaZe Clan;\\ \textbf{Context:}  February 18, 2018. FaZe Clan FaZe Clan (formerly FaZe Sniping) is an American esports and entertainment organization that competes in various video game tournaments. The organization was founded as a gaming clan on YouTube by players known as Housecat, ClipZ, and Resistance in 2010, who all created "trickshot" videos for the video game "". In 2012, with the release of "", the organization decided to expand into competitive play. In 2016, a new era for FaZe began when the organization bought a "" professional team. This moment marked the beginning of FaZe Clan expanding into various esports. This movement is;\\ \textbf{Title:}  FaZe Clan;\\ \textbf{Context:}  FaZe Clan FaZe Clan (formerly FaZe Sniping) is an American esports and entertainment organization that competes in various video game tournaments. The organization was founded as a gaming clan on YouTube by players known as Housecat, ClipZ, and Resistance in 2010, who all created "trickshot" videos for the video game "". In 2012, with the release of "", the organization decided to expand into competitive play. In 2016, a new era for FaZe began when the organization bought a "" professional team. This moment marked the beginning of FaZe Clan expanding into various esports. This movement is referred to as;\\ \textbf{Title:}  FaZe Clan;\\ \textbf{Context:}  Duty" community has been frustrated with the way FaZe Clan has been/not been involved in the game in which it was founded. Most of the other FaZe members have changed their video topics and mostly do vlog videos. This sparked FaZe Clan to organize the highly successful FaZe Bootcamp on the release of "". FaZe trickshotters known as Kitty, Dirty, Bloo, GwidT and Replays contributed to a week dedicated to creating "Call of Duty" content. FaZe's first roster—consisting of players named Heist, Folsom, Secretly, and Sham— was created to compete at the 2013 MLG Winter Championship. Their next roster—consisting of;\\ \textbf{Title:}  F Is for Family;\\ \textbf{Context:}  F Is for Family F is for Family is an American adult animated sitcom created by Bill Burr and Michael Price and produced by Gaumont International Television and Vince Vaughn's Wild West Television. The show premiered on December 18, 2015, to generally favorable reviews. Season 2 premiered May 30, 2017. On June 28, 2017, the show was renewed for a third season. On July 1, 2018, Burr confirmed season three. On November 30, the third season was released. The series was announced in October 2014 as part of a partnership between Netflix, Gaumont International Television, and Wild West Television.; \\ \textbf{What is the answer of the following question:}\\
who has the most subscribers in faze clan \\
\hline
\textbf{GTR Prediction}: Tfue \\
\hline
\hline
\textbf{PSR Context} \\
\hline
\textbf{Given the passage titles and contexts below:} \\
\textbf{Title:} FaZe Clan;\\ \textbf{Context:}  Duty" community has been frustrated with the way FaZe Clan has been/not been involved in the game in which it was founded. Most of the other FaZe members have changed their video topics and mostly do vlog videos. This sparked FaZe Clan to organize the highly successful FaZe Bootcamp on the release of "". FaZe trickshotters known as Kitty, Dirty, Bloo, GwidT and Replays contributed to a week dedicated to creating "Call of Duty" content. FaZe's first roster—consisting of players named Heist, Folsom, Secretly, and Sham— was created to compete at the 2013 MLG Winter Championship. Their next roster—consisting of;\\ \textbf{Title:}  FaZe Clan;\\ \textbf{Context:}  "FaZe 2.0" by FaZe members and fans. The FaZe "CS:GO" team went on to become one of the most successful rosters for the 2017/2018 seasons. FaZe Clan is the most popular esports organization in the world, based on the organization's social media following. As of November 28, 2018, FaZe Clan and its members together have 82 million YouTube subscribers, 11.2 billion YouTube views, 11.3 million Twitch followers, 130 million Twitch views, 43.1 million Twitter followers, 45.8 million Instagram followers, 2.8 million Facebook likes and followers. FaZe Clan has made \$6,148,290.91 from esport tournament prize pools alone. FaZe Clan started on;\\ \textbf{Title:}  FaZe Clan;\\ \textbf{Context:}  FaZe Clan FaZe Clan (formerly FaZe Sniping) is an American esports and entertainment organization that competes in various video game tournaments. The organization was founded as a gaming clan on YouTube by players known as Housecat, ClipZ, and Resistance in 2010, who all created "trickshot" videos for the video game "". In 2012, with the release of "", the organization decided to expand into competitive play. In 2016, a new era for FaZe began when the organization bought a "" professional team. This moment marked the beginning of FaZe Clan expanding into various esports. This movement is referred to as;\\ \textbf{Title:}  F Is for Family;\\ \textbf{Context:}  F Is for Family F is for Family is an American adult animated sitcom created by Bill Burr and Michael Price and produced by Gaumont International Television and Vince Vaughn's Wild West Television. The show premiered on December 18, 2015, to generally favorable reviews. Season 2 premiered May 30, 2017. On June 28, 2017, the show was renewed for a third season. On July 1, 2018, Burr confirmed season three. On November 30, the third season was released. The series was announced in October 2014 as part of a partnership between Netflix, Gaumont International Television, and Wild West Television.;\\ \textbf{Title:}  FaZe Clan;\\ \textbf{Context:}  February 18, 2018. FaZe Clan FaZe Clan (formerly FaZe Sniping) is an American esports and entertainment organization that competes in various video game tournaments. The organization was founded as a gaming clan on YouTube by players known as Housecat, ClipZ, and Resistance in 2010, who all created "trickshot" videos for the video game "". In 2012, with the release of "", the organization decided to expand into competitive play. In 2016, a new era for FaZe began when the organization bought a "" professional team. This moment marked the beginning of FaZe Clan expanding into various esports. This movement is; \\
\textbf{What is the answer of the following question: }\\
who has the most subscribers in faze clan \\
\hline
\textbf{PSR Prediction}: Tfue \\
\hline
\hline
\textbf{BGM Context} \\
\hline
\textbf{What is the answer of the following question: }\\
who has the most subscribers in faze clan \\
\hline
\textbf{BGM Prediction}: FaZe Rug\\
\toprule
\caption{Cases comparisons for GTR, PSR and BGM from NQ dataset.}
\label{ap_tab.case}
\end{longtable}

\twocolumn

\end{document}